\documentclass[10pt,twocolumn,letterpaper]{article}

\usepackage[pagenumbers]{wacv} 

\usepackage{graphicx}
\usepackage{amsmath}
\usepackage{amssymb}
\usepackage{booktabs}
\usepackage{multirow}

\usepackage[pagebackref,breaklinks,colorlinks]{hyperref}

\usepackage[capitalize]{cleveref}
\crefname{section}{Sec.}{Secs.}
\Crefname{section}{Section}{Sections}
\Crefname{table}{Table}{Tables}
\crefname{table}{Tab.}{Tabs.}

\def\wacvPaperID{1843} 
\def\confName{WACV}
\def\confYear{2025}

\begin{document}

\title{RopeBEV: A Multi-Camera Roadside Perception Network in Bird's-Eye-View}

\author
{ 
Jinrang Jia\thanks{Equal Contribution.}
\quad
Guangqi Yi$^*$
\quad
Yifeng Shi\thanks{Corresponding Author.}
\\
Baidu Inc.
\\
\tt\small\{jiajinrang, yiguangqi, shiyifeng\}@baidu.com
}
\maketitle

\begin{abstract}
Multi-camera perception methods in Bird’s-Eye-View (BEV) have gained wide application in autonomous driving. However, due to the differences between roadside and vehicle-side scenarios, there currently lacks a multi-camera BEV solution in roadside. This paper systematically analyzes the key challenges in multi-camera BEV perception for roadside scenarios compared to vehicle-side. These challenges include the diversity in camera poses, the uncertainty in Camera numbers, the sparsity in perception regions, and the ambiguity in orientation angles. In response, we introduce RopeBEV, the first dense multi-camera BEV approach. RopeBEV introduces BEV augmentation to address the training balance issues caused by diverse camera poses. By incorporating CamMask and ROIMask (Region of Interest Mask), it supports variable camera numbers and sparse perception, respectively. Finally, camera rotation embedding is utilized to resolve orientation ambiguity. Our method ranks 1st on the real-world highway dataset RoScenes and demonstrates its practical value on a private urban dataset that covers more than 50 intersections and 600 cameras.
\end{abstract}

\section{Introduction}
\label{sec:intro}

With the rapid development of the intelligent transportation \cite{Rukhovich_2022_WACV, nwad121, danet, DUSA_2023_ACMMM, fan2021embracingsinglestride3d, Borse_2023_WACV, Fadadu_2022_WACV}, the use of roadside cameras for traffic perception has garnered increasing attention. Roadside cameras are typically mounted on poles, positioned 6-15 meters above the ground \cite{ye2022rope3d, zhu2024rosceneslargescalemultiview3d}. To achieve collaborative perception, multiple roadside cameras are usually required in an intersection or corridor scenario.

Collaborative perception can be categorized into three classes based on the phase at which multi-sensor fusion occurs: early fusion (fusion at the raw data level), intermediate fusion (fusion at the feature level) \cite{Chen_2023_ICCV, Qiao_2023_WACV, lu2023robust, Where2comm, xu2022v2xvit}, and late fusion (fusion at the perception result level) \cite{Yu_2022_CVPR}. Currently, most roadside collaborative perception systems adopt the late fusion approach, where individual camera perception is first performed \cite{10550788, peng2022did, jia2023monouni, fan2023calibrationfreebevrepresentationinfrastructure, 10186723}, followed by the fusion of the results from each camera \cite{Yu_2022_CVPR}. Although the late fusion approach has the advantage of being easier to implement, it often faces challenges such as object splitting or positional jitter when objects span across multiple cameras or when occlusions occur.

The multi-camera fusion approach from the BEV perspective \cite{Yang2022BEVFormerVA, philion2020lift, li2022bevdepth, Peng_2023_WACV, Chen_2024_WACV} has been widely used in vehicle-side perception and has been thoroughly validated for its stability compared to late fusion methods. BEV methods can be classified into sparse \cite{detr3d, liu2023sparsebevhighperformancesparse3d, liu2022petr, Sparse4D, lin2023sparse4d} and dense BEV \cite {yang2023bevformer, Zhang_2023_ICCV} approaches based on their modeling techniques. Compared to sparse BEV methods, dense approaches can explicitly generate BEV features for the entire space, further supporting downstream tasks such as lane segmentation and scene mapping. However, due to the differences between roadside and vehicle-side multi-camera perception scenarios, dense BEV methods designed for vehicles cannot be directly applied to roadside environments.

This paper begins with a systematic analysis of the differences between roadside and vehicle-side multi-camera perception scenarios, as listed below: (1) \textbf{Diversity in Camera Poses}. In vehicle-side scenarios, the relative pose of the cameras remains constant regardless of the vehicle’s location. However, in roadside scenarios, the setup of cameras varies significantly from one location to another, leading to differing poses between cameras at each site. (2) \textbf{Uncertainty in Camera Numbers}. For a specific vehicle model, the number of cameras is fixed, and this number only changes in extreme cases, such as camera malfunction. In contrast, the number of cameras involved in collaborative perception in roadside scenarios varies widely across different sites. (3) \textbf{Sparsity in Perception Regions}. Roadside cameras are installed on poles at a height of 6 to 15 meters, providing a much broader field of view. Roadside images may contain regions where obstacles appear infrequently or not at all, leading to ineffective perception and resource wastage. (4) \textbf{Ambiguity in Orientation Angles}. In vehicle-side systems, the BEV coordinate system is ego-centric, with its origin fixed to the vehicle’s coordinate system. In contrast, roadside scenarios use a non-ego-centric coordinate system where the origin of the BEV coordinate system is not fixed. This creates potential ambiguity in orientation angle of objects, theoretically leading to a 1-to-X problem.

In response to the aforementioned differences in scenarios, this paper proposes a dense BEV perception method for roadside environments, named RopeBEV. RopeBEV introduces a BEV augmentation technique to address the issue of unbalanced training of learnable queries caused by the diversity in roadside camera poses, enabling each query to extract and transform features from any 2D to 3D perspective. Additionally, RopeBEV incorporates two masking mechanisms, CamMask and ROIMask, which allow the framework to support an arbitrary number of cameras and achieve more efficient feature utilization. Finally, camera rotation embedding is introduced to resolve the ambiguity of object orientation angles in the BEV space.

To the best of our knowledge, RopeBEV is the first dense BEV method designed for roadside scenarios. It has achieved the top ranking on the real-world roadside highway dataset RoScenes \cite{zhu2024rosceneslargescalemultiview3d}. Additionally, to validate the effectiveness of RopeBEV, we trained our model on a large real-world urban dataset that covers over 50 intersections and 600 cameras. The results demonstrate its superiority in industrial applicability and potential for real-world deployment. Our contributions can be summarized as:
\begin{itemize}
\item We systematically analyze the differences between roadside and vehicle-side multi-camera perception scenarios, including diversity in camera poses, uncertainty in camera numbers, sparsity in perception regions, and ambiguity in orientation angles.
\item We propose the first dense BEV method for roadside scenarios, RopeBEV, which incorporates improvements specifically designed to address these four differences: BEV Augmentation, CamMask, ROIMask, and Camera Rotation Embedding.
\item RopeBEV ranks 1st on the real-world highway dataset RoScenes and demonstrates its industrial applicability on a large-scale private urban dataset with over 50 intersections and 600 cameras.
\end{itemize}
\section{RELATED WORK}
\label{sec:related_work}

\subsection{Roadside Camera-based Perception} 
Over the past few years, research on roadside camera-based perception has primarily focused on monocular 3D detection, leading to the release of several benchmarks for monocular 3D detection in roadside scenarios. Rope3D \cite{ye2022rope3d}, was the first real-world dataset specifically designed for roadside monocular 3D detection. DAIR-V2X \cite{Yu_2022_CVPR} added vehicle-side data, and V2X-Seq \cite{v2x-seq} further extended it to a sequential dataset. Building on these datasets, numerous methods for roadside monocular 3D detection have been proposed. BEVHeight \cite{yang2023bevheight} and BEVHeight++ \cite{yang2023bevheightrobustvisualcentric} improved 3D detection accuracy by regressing the object's height relative to the ground instead of directly regressing depth. MonoUNI \cite{jia2023monouni} introduced the concept of normalized depth to mitigate ambiguities caused by focal length and pitch angle. MonoGAE \cite{yang2023monogaeroadsidemonocular3d} incorporated prior knowledge of the ground plane. BEVSpread \cite{wang2024bevspreadspreadvoxelpooling} optimized the generation of BEV features of single camera and further enhance performance. To promote the development of collaborative perception, RoScenes \cite{zhu2024rosceneslargescalemultiview3d} created the first roadside multi-camera 3D detection dataset for high-speed scenarios, with each scene containing 6 to 12 cameras covering a perception range of 800x80 meters. RCooper \cite{hao2024rcooper} released a roadside multi-camera dataset, covering intersection and corridor scenarios.

\subsection{Multi-Camera BEV Perception} 
Multi-camera BEV perception can be categorized into two main approaches: sparse BEV and dense BEV. Sparse BEV methods, represented by DETR3D \cite{detr3d} and PETR \cite{liu2022petr}, are transformer-based approaches. They primarily work by combining 3D position embedding with 2D features to directly generate fused features with multi-view positional information. These features are then processed through a transformer-based decoder and subsequent task heads to achieve multi-view perception, without the need to explicitly generate dense BEV features. On the other hand, dense BEV methods, exemplified by algorithms such as BEVDet \cite{huang2021bevdet}, BEVFormer \cite{li2022bevformer}, and FastBEV \cite{li2023fast}, take a different approach. Multi-camera images are first used to extract 2D features from each camera's perspective. These 2D features are then fused into a dense BEV feature map using intrinsic and extrinsic camera parameters. This dense feature map serves as the basis for subsequent perception tasks. The dense approach is particularly advantageous for implementing temporal fusion and further supports tasks such as lane perception and scene mapping. Building on the basic framework of vehicle-side, this paper introduces the first multi-camera dense BEV method specifically designed for roadside scenarios.

\section{Method}

In this section, we first introduce the overall structure of RopeBEV. Then, we provide a detailed analysis of the differences between roadside and vehicle-side perception scenarios, including the diversity in camera poses, the uncertainty in camera numbers, the sparsity in perception regions, and the ambiguity in orientation angles. Finally, the improvements introduced in RopeBEV to address these challenges are discussed.

\begin{figure*}
\centering
\includegraphics[width=170mm]{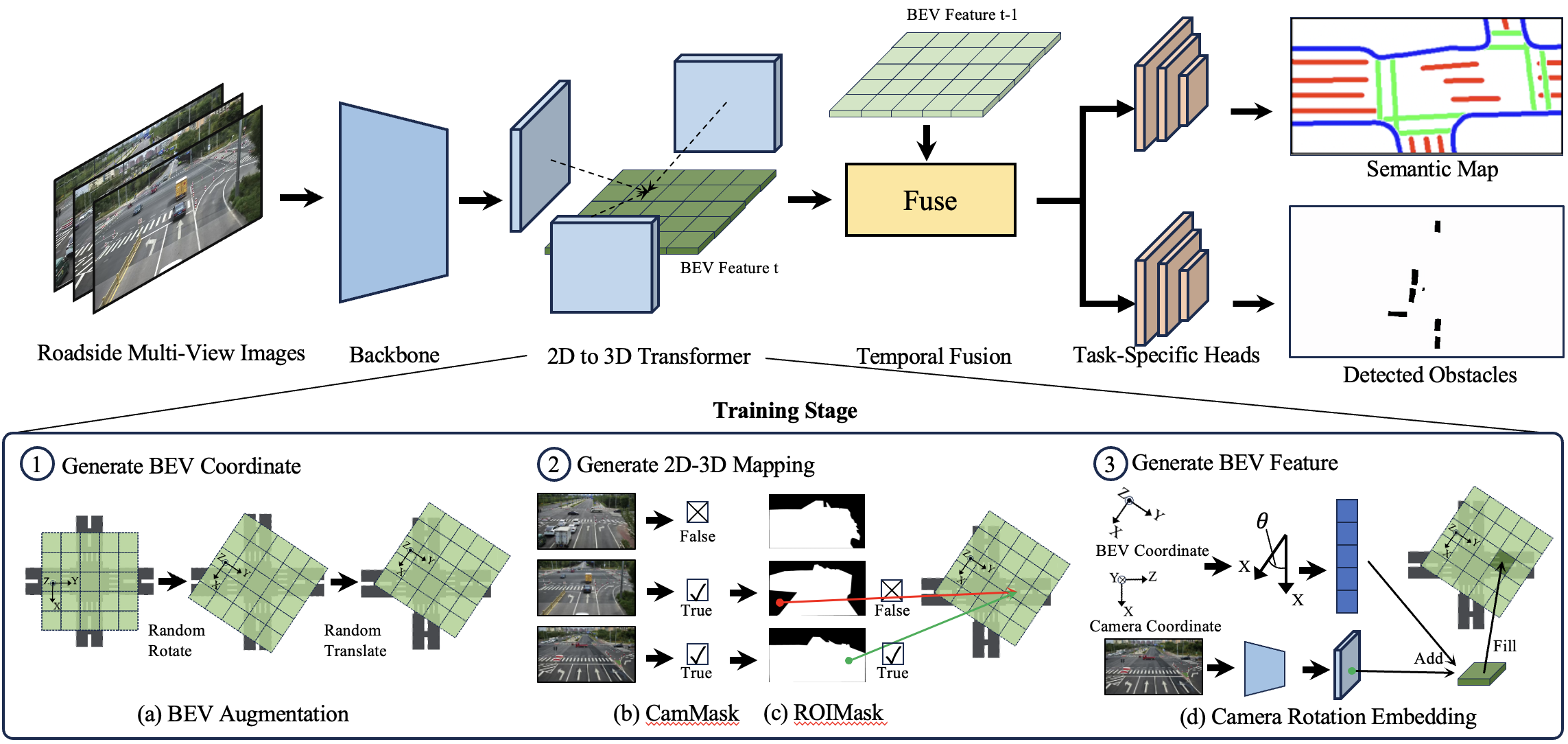}
\caption{\textbf{An overview of the RopeBEV framework.} The overall network follows a typical dense BEV architecture, which includes a backbone, a 2D-to-3D transformer, a temporal fusion module, and several task-specific heads. Considering the characteristics of roadside scenarios, RopeBEV introduces improvements in the 2D-to-3D transformer. The 2D-to-3D transformer can be divided into three stages: (1) \textbf{Generate BEV Coordinate}. In this stage, RopeBEV introduces BEV coordinate system data augmentation to address the training imbalance caused by the diversity of roadside camera poses. (2) \textbf{Generate 2D-3D Mapping}. Here, RopeBEV incorporates CamMask and ROIMask mechanisms to support customizable camera numbers and perception regions. (3) \textbf{Generate BEV Feature}. In this stage, RopeBEV integrates Camera Rotation Embedding into the features of single cameras to resolve orientation angle ambiguities.}
\label{overview}
\end{figure*}

\subsection{Overall} 
As shown in Figure \ref{overview}, the overall framework is built on the BEVFormer \cite{li2022bevformer} architecture and consists of four main modules: (1) backbone, (2) 2D to 3D transformer, (3) temporal fusion and (4) task-specific heads. The backbone takes several images $I^t = \{I^t_1, I^t_2, ..., I^t_N \}$ captured at time $t$ as inputs and compute the deep 2D features $F^t = \{F^t_1, F^t_2, ..., F^t_N \}$, where $N$ is the total camera number. The 2D to 3D transformer module integrates and transforms $F^t$ into a global BEV feature $F^t_{BEV}$. In the temporal fusion module, $F^t_{BEV}$ is further fused with features $F^{t-1}_{BEV}$ from previous time steps. Finally, the temporally fused feature $F^t_{BEV}$ is passed into various downstream heads to produce outputs for different tasks. 

\subsection{Diversity in Camera Poses}

\begin{figure}[t]
\begin{center}
\includegraphics[width=1.0\linewidth]{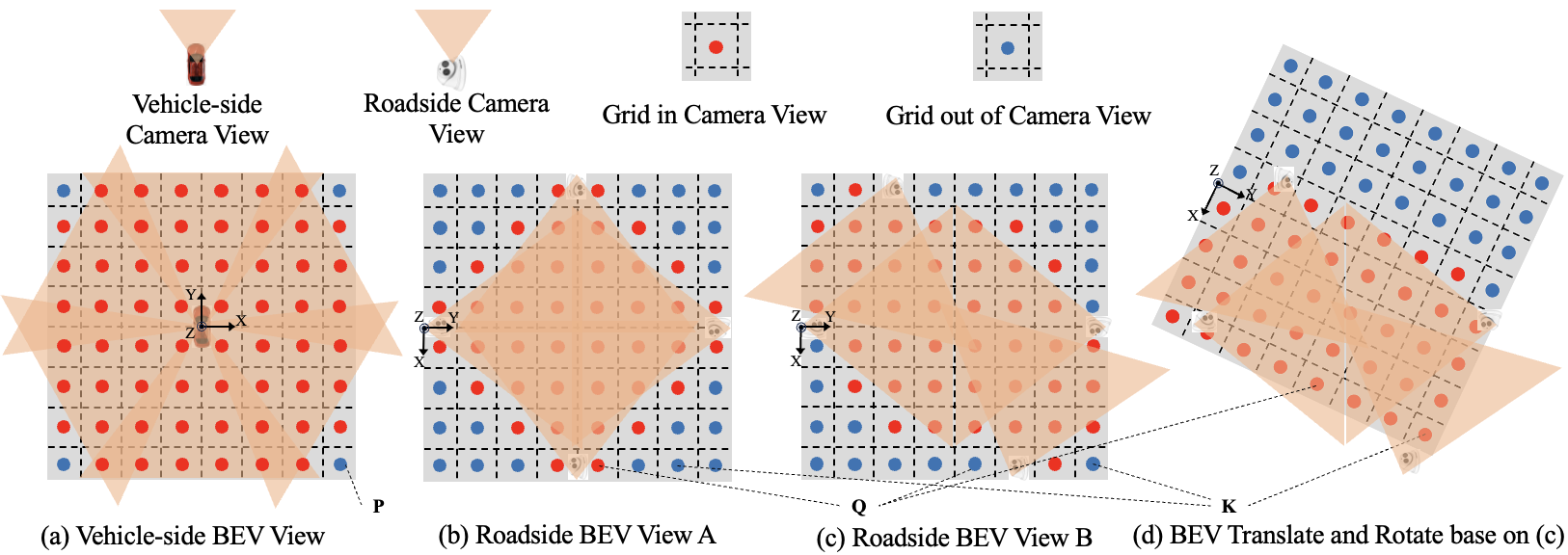}
\end{center}
\caption{\textbf{Camera Views on vehicle-side and different roadside scenes.} (a) illustrates the Camera View in a vehicle-side scenario. Regardless of where the vehicle travels, this view remains unchanged. Although Grid $P$ is never trained, it is also never utilized. (b) and (c) depict Camera Views from two different roadside scenarios. Due to the variability of real-world scenes, the Camera Views differ, leading to an imbalance in training. For instance, Grid $Q$ is trained in (b) but not in (c), while Grid $K$ is not trained in either (b) or (c). However, both Grid $Q$ and $K$ might still be used in other scenes, which could result in performance issues due to insufficient training. (d) demonstrates the application of BEV coordinate system translation and rotation for data augmentation in scenario (c). This augmentation allows Grids $Q$ and $K$ to be trained, addressing the training imbalance and ensuring that these grids are better prepared for use in various roadside scenarios.}
\label{camera_pose}
\end{figure}

The 2D to 3D transformer module is the most critical component in BEV perception and can generally be divided into three stages: (1) Generate BEV Coordinate, (2) Generate 2D-3D Mapping, and (3) Generate BEV feature. In the first stage, the generation of the BEV coordinate system is related to the layout of cameras. As shown in Figure \ref{camera_pose} (a), for autonomous vehicles, the BEV coordinate system is typically equivalent to the LiDAR coordinate system or the vehicle body coordinate system. For a given vehicle model, the poses of the camera sensors relative to the BEV coordinate system are fixed, meaning that the mapping between 2D pixels and 3D BEV grids remains constant. This fixed mapping significantly reduces the learning difficulty for feature extractors in 2D-to-3D transformer modules that utilize spatially-aware learnable parameters (e.g., BEVFormer \cite{li2022bevformer}). In such cases, a specific feature extractor only needs to focus on specific patterns. Moreover, grids outside the camera's field of view (such as the blue grid $P$ in Figure \ref{camera_pose} (a)) are never trained, but this does not negatively impact the autonomous vehicle's capabilities because these areas will never be visible to the cameras, regardless of where the vehicle travels. In fact, for most autonomous vehicles equipped with multiple surround-view cameras, there are no blind spots similar to point $P$ in Figure \ref{camera_pose} (a). 

However, for roadside perception, the situation is quite different. Since real-world scenes vary widely, the camera layout strategies also differ from one location to another. Consequently, the poses of cameras relative to the BEV coordinate system vary across different scenes. This variation leads to an imbalance in training the feature extractors. As illustrated in Figures \ref{camera_pose} (b) and (c), a specific grid $Q$ might be trained in one scene (Figure \ref{camera_pose} (b)) but not in another (Figure \ref{camera_pose} (c)). This imbalance results in some feature extractors being undertrained or even not trained at all (such as grid $K$). Unlike in vehicle-side scenarios, these undertrained extractors may be used in new scenes, potentially leading to performance issues. Additionally, the learning difficulty for feature extractors in roadside scenarios is further increased because the patterns that a specific extractor needs to learn  differ significantly across various scenes. 

RopeBEV addresses this issue by employing BEV data augmentation. As shown in Figure \ref{camera_pose} (d), during the training phase, the BEV coordinate system is randomly translated and rotated, ensuring that each feature extractor within the BEV grid is trained more uniformly. This approach mitigates the problem of undertrained feature extractors. However, the issue of feature extractors needing to adapt to non-fixed patterns in roadside scenarios still persists. It is worth noting that for methods without spatially-aware learnable parameters, such as FastBEV \cite{li2023fast}, there is no need to introduce BEV augmentation.

\subsection{Uncertainty in Camera Numbers}
For autonomous vehicles, the layout of camera sensors is fixed, and the number of cameras is also predetermined, with changes only occurring in extreme cases like camera damage. In contrast, in roadside scenarios, the number of cameras varies naturally across different scenes due to the diverse real-world road conditions. This variability necessitates that multi-camera perception networks support the training with a dynamic number of cameras. As shown in Figure \ref{overview} (b), RopeBEV addresses this by introducing a CamMask in the 2D to 3D transformer module. Specifically, during training, cameras for which the CamMask is set to \textit{False} are excluded from the 2D to 3D mapping process, meaning that their features do not contribute to the subsequent BEV feature construction. The CamMask ensures that at least one and up to a maximum of $N$ cameras are active, where $N$ represents the maximum number of supported cameras. During inference, CamMask is configured according to the actual camera number. This approach allows RopeBEV to effectively handle scenarios with varying camera configurations, ensuring robust perception across diverse roadside environments.

\begin{figure}[t]
\begin{center}
\includegraphics[width=1.0\linewidth]{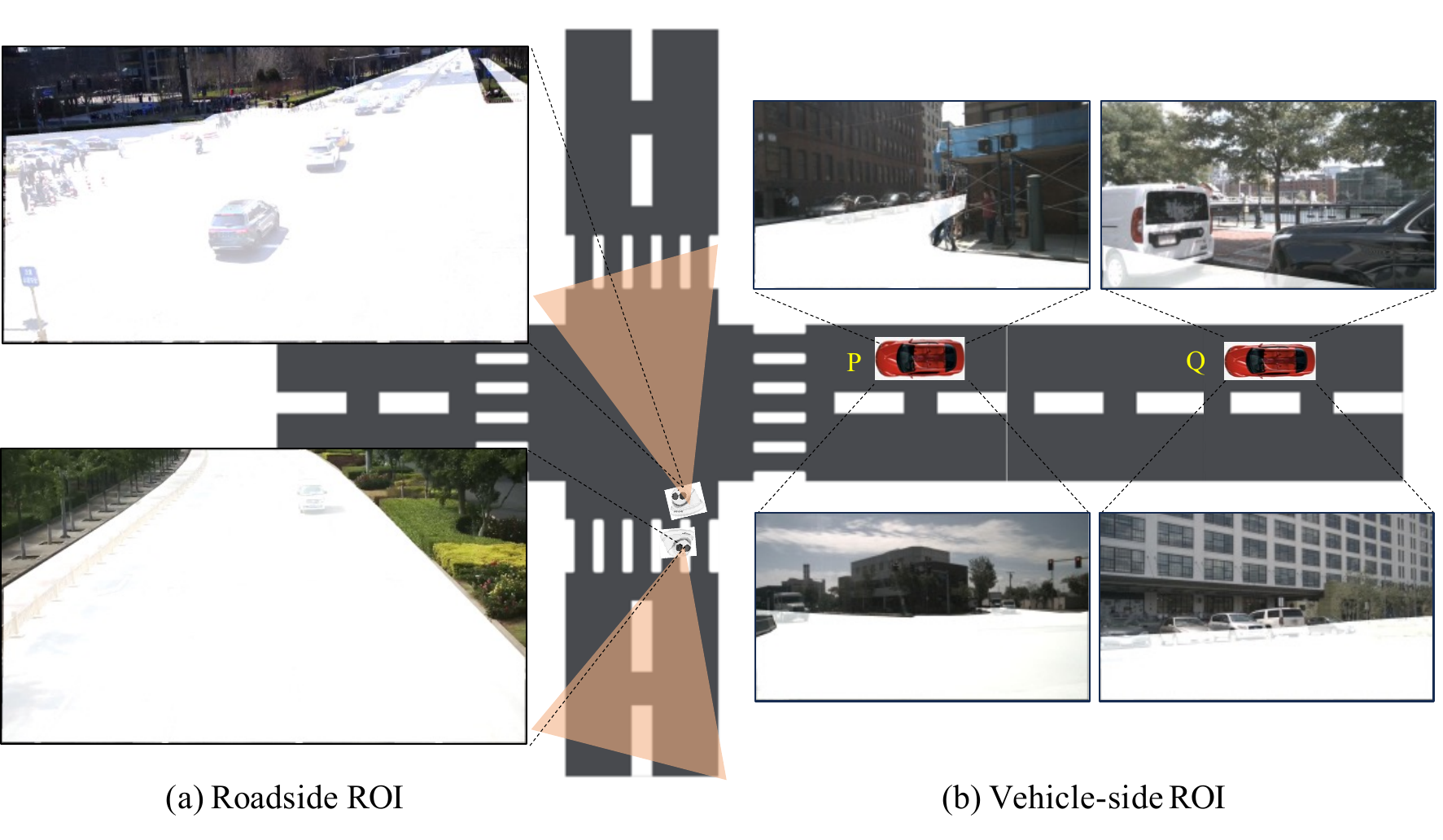}
\end{center}
\caption{\textbf{ROIs on vehicle-side and roadside scenes.} White regions on images are ROIs. Since roadside cameras are stationary, their ROIs are also fixed. However, because vehicles are in motion (from $Q$ to $P$), their ROIs vary as the vehicle's position changes. This distinction enables the use of ROIMask in roadside scenarios to filter out irrelevant perception areas, a method that cannot be applied to vehicle-side cameras.}
\label{roi}
\end{figure}

\subsection{Sparsity Perception Regions}
As illustrated in Figure \ref{roi} (a), roadside cameras are mounted on poles at heights ranging from 6 to 15 meters, resulting in a wide field of view. This broad perspective often captures areas where obstacles appear infrequently or not at all, leading to inefficient perception and wasted resources. As shown in Figure \ref{overview} (c), RopeBEV addresses this issue by incorporating a 2D Region of Interest (ROI) for each roadside camera and introducing an ROIMask filtering mechanism during training to support the customization of effective perception areas. When establishing the 2D-3D mapping relationship, image features outside the ROI are excluded from the mapping process, meaning that these irrelevant features do not contribute to the BEV Feature construction. Although the resulting BEV feature is dense, every part of it pertains to valuable scenes, effectively excluding non-essential features such as the sky or non-road areas. This optimization fundamentally benefits from the fixed positioning of roadside cameras, enabling the algorithm to support fine-grained perception through the use of ROIMask. It also allows for the customization of the perception area according to downstream requirements. As shown in Figure \ref{roi} (b), although vehicle-side cameras also have areas of irrelevant perception, they cannot achieve fine-grained perception due to the constant movement of the vehicle. The fine-grained perception in roadside scenarios also contributes to reducing computational resource consumption.

\begin{figure}[t]
\begin{center}
\includegraphics[width=1.0\linewidth]{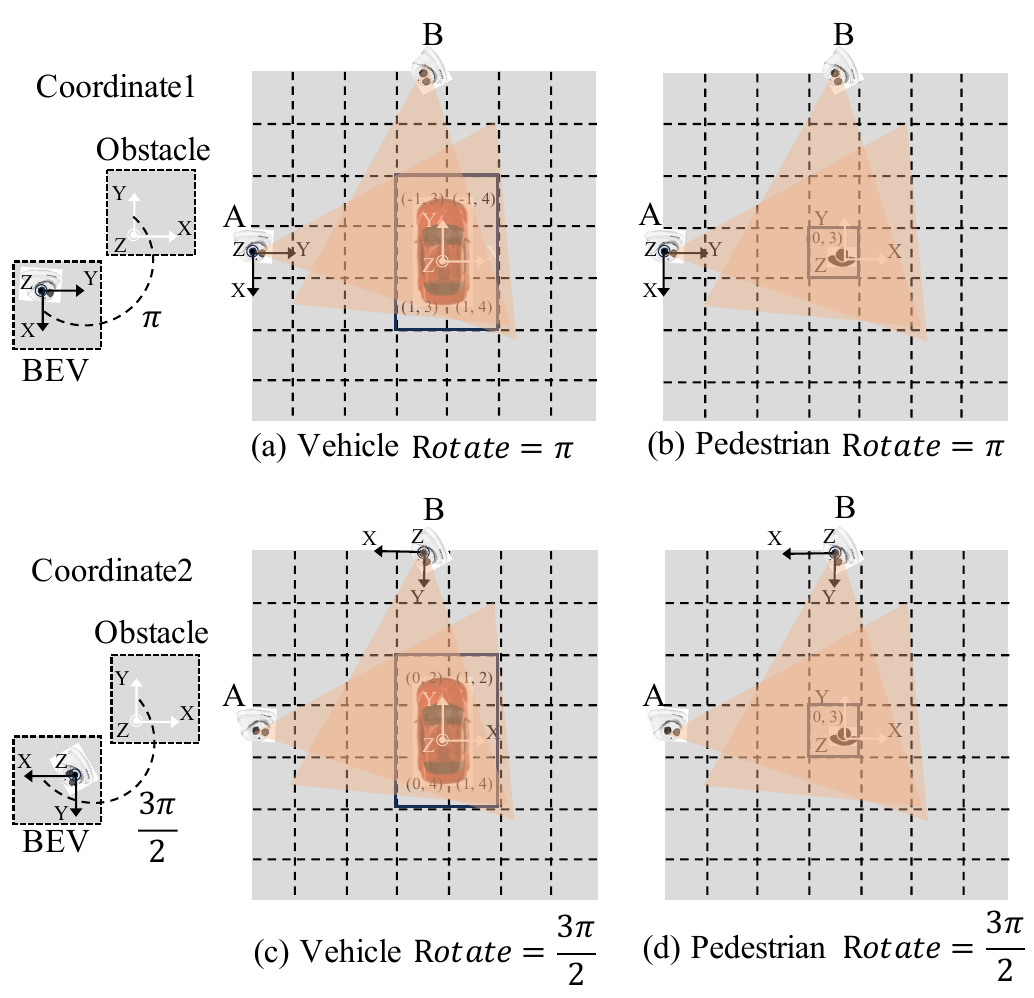}
\end{center}
\caption{\textbf{Ambiguity in Orientation Angles.} The camera deployment schemes in (a), (b), (c), and (d) are identical. In (a) and (c), the obstacle is a vehicle occupying multiple grids, whereas in (b) and (d), the obstacle is a pedestrian occupying only a single grid. The BEV coordinate system in (a) and (b) is centered at Camera A, with the Y-axis pointing to the right. As shown in the left-side schematic, the orientation angle of the obstacle is $\pi$. In (c) and (d), the BEV coordinate system is centered at Camera B, with the Y-axis pointing downward, and the obstacle's orientation angle is $\frac{3\pi}{2}$. When the BEV coordinate system shifts from (a) to (c), the BEV feature of the vehicle remains unchanged, but the 3D position encoding changes, resulting in a change in the orientation angle without ambiguity. However, when the BEV coordinate system shifts from (b) to (d), both the BEV feature and the 3D position encoding of the pedestrian remain unchanged, yet the orientation angle changes, leading to ambiguity.}
\label{rotate}
\end{figure}

\subsection{Ambiguity in Orientation Angles}
In 3D detection, the model needs to predict the orientation angle of each obstacle based on the BEV features. The orientation angle of an obstacle is defined as the angle between the obstacle's orientation (the Y-axis of the white obstacle coordinate system in Figure \ref{rotate}) and the horizontal axis of the coordinate system it is located in (the X-axis of the black BEV coordinate system in Figure \ref{rotate}), measured around the vertical axis (the Z-axis of the BEV coordinate system in Figure \ref{rotate}).  

As this definition implies, the orientation angle is relative to the BEV coordinate system where the obstacle is situated. When the BEV coordinate system rotates, the orientation angle of the obstacle should also change accordingly. For instance, as described in the transitions from Figure \ref{rotate} (a) to (c) or from (b) to (d), the obstacle (vehicle in (a)(c) or pedestrian in (b)(d)), along with cameras A and B, remains stationary, while the BEV coordinate system shifts from being centered on camera A to being centered on camera B, rotating $\frac{\pi}{2}$ in the process. Despite the BEV feature remaining unchanged between (a) and (c) (or (b) and (d)) due to the static positions of the obstacle and cameras, the orientation angle of the obstacle changes. This results in a 1-to-X ambiguity: the input to the model stays the same, but the output (orientation angle) changes.

However, this ambiguity is typically resolved when obstacles span multiple BEV grids. This is because BEV features often include 3D positional embedding, which provides enough structural information to describe the obstacle's orientation across several BEV grids. The 3D positional embedding, which changes with the BEV coordinate system's rotation, helps maintain consistent orientation predictions under the grid's rotation. But when an obstacle occupies only a single BEV grid, there's insufficient information to fully describe the orientation, and in some cases, the 3D positional embedding might not change at all (as in Figures \ref{rotate} (b) and (d)). This leads to ambiguity, where the model need to predict different orientation angles based on same input.

The core of the orientation angle ambiguity issue arises from the non-ego-centric nature of roadside perception, where the BEV coordinate system is not fixed, especially with the BEV augmentation. Unlike roadside scenarios, in vehicle-side perception systems, the BEV coordinate system is fixed relative to the vehicle, and the camera positions are consistent, so this issue does not occur. To address the orientation angle ambiguity, one could increase the density of the BEV grid so that each obstacle spans multiple grids, providing enough structural information for accurate orientation prediction. However, denser grids lead to higher computation and latency. As illustrated in Figure \ref{overview} (c), RopeBEV resolves this ambiguity by explicitly adding the camera's rotation angle as an embedding to the single-camera 2D feature. This approach supplements the missing information without needing to increase grid density. Given $F=\{F_1, F_2,..., F_N\}$, where each $F_n \in\mathbb{R}^{c\times h \times w}$ represents the 2D feature of the $n$-th camera. $\theta=\{\theta_1, \theta_2,..., \theta_N\}$, where each $\theta_n \in\mathbb{R}^1$ represents the angle between the orientation of the $n$-th camera and the BEV coordinate system's orientation. To avoid the periodicity issues of directly using numerical embeddings, we use $sine$ and $cosine$ encoding for the angles \cite{vaswani2023attentionneed}. The process of adding the camera rotation embedding can be expressed as:

\begin{equation}
\begin{aligned}
    F^{'}_{i} = F_{i} + Expand(Embed([sin(\theta_n), cos(\theta_n)]))
\end{aligned}
\end{equation}
where $Embed()$ is embedding function that maps the rotation angle $[sin(\theta_n), cos(\theta_n)]$ into the feature space $\mathbb{R}^{c}$. $Expand()$ is the expanding function which makes the embedding size same with feature $\mathbb{R}^{c\times h \times w}$.

\section{Experiments}

In this section, we first introduce two multi-camera roadside datasets and the implementation details of our RopeBEV. Then, we compare our proposed RopeBEV with state-of-the-art methods. Subsequently, in the ablation study section, we quantitatively discuss the effects of BEV Augmentation and Camera Rotation Embedding. Finally, we present the qualitative effects of CamMask and ROIMask through visualized results, showcasing the overall perception capabilities of RopeBEV.

\subsection{Datasets}
\textbf{RoScenes.} 
Roscenes \cite{zhu2024rosceneslargescalemultiview3d} is a large-scale multi-view roadside perception dataset which  includes significantly large perception area, full scene coverage and crowded traffic. It contains $1.30$ million images from roadside mounted cameras in 14 highway scenes. Over $21$ million 3D boxes are annotated within 4 classes: Car, Van, Bus and Truck. We use the official training and validation set splits. For the validation set, by sorting all clips via clip-level multi-view occlusion, we label the first 50\% of the data as Easy and the remaining 50\% as Hard. The evaluation main metrics include NuScenes Detection Score (NDS), mean Average Precision (mAP), mean Average Translation Error (mATE), mean Average Scale Error (mASE) and mean Average Orientation Error (mAOE).

\begin{figure}[t]
\begin{center}
\includegraphics[width=1.0\linewidth]{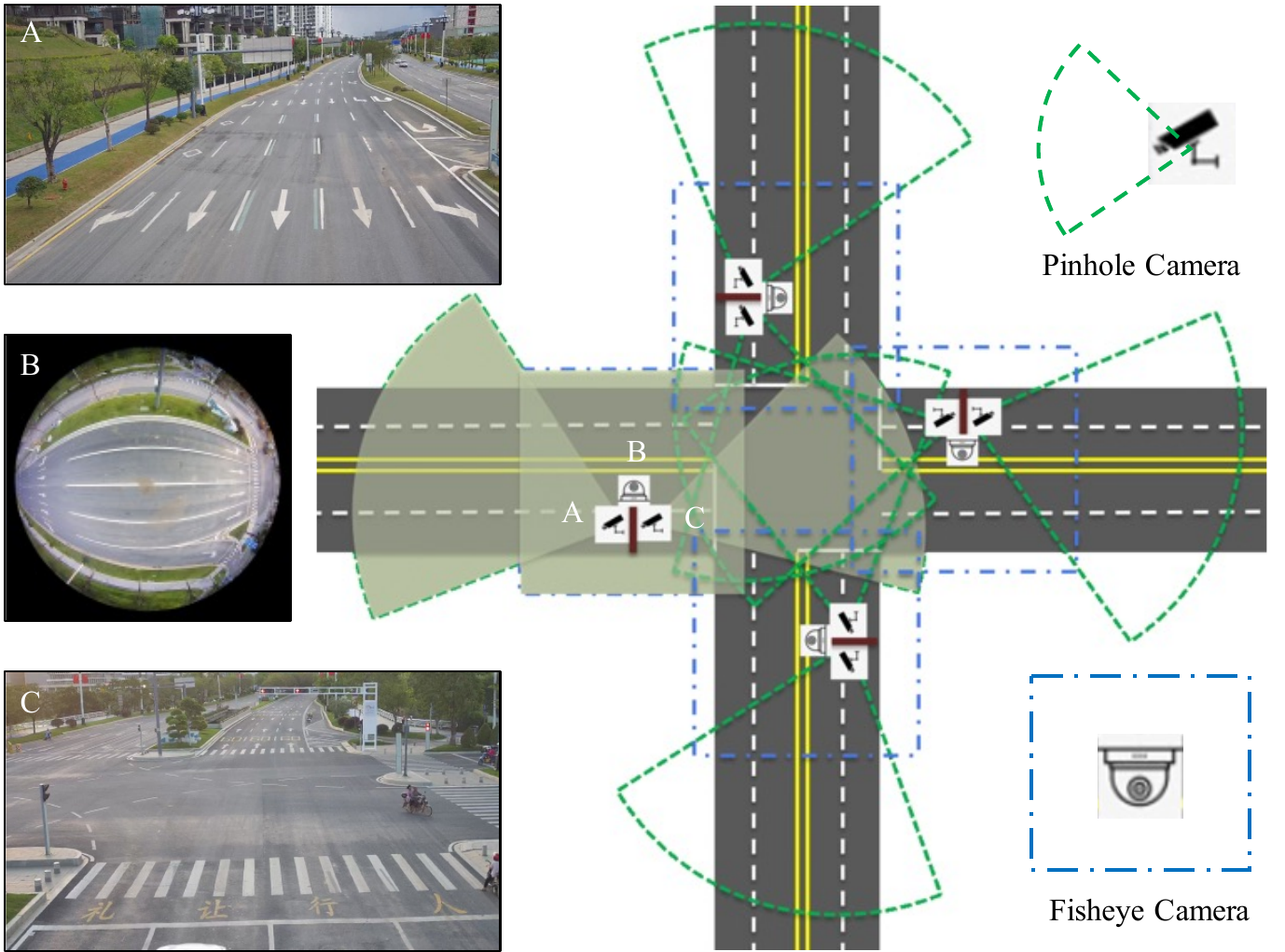}
\end{center}
\caption{\textbf{Camera deployment of private dataset.} Each intersection has 8 pinhole and 4 fisheye cameras.}
\label{private}
\end{figure}

\textbf{Private Dataset.}
Since RoScenes is a high-speed scenario dataset and there is currently a lack of large-scale, multi-camera annotated datasets for urban scenes in academia\footnote{There was no download permission for Rcooper \cite{hao2024rcooper} when the paper was written.}, we conducted an extensive evaluation of RopeBEV on a private dataset that includes more than 50 urban intersections and 600 cameras. As shown in Figure \ref{private}, each intersection's data consists of 8 pinhole cameras and 4 fisheye cameras, with an average recording duration of over 1 minute per intersection.The image data is collected at a frequency of 13 Hz and is semi-automatically annotated. This private dataset totally provides over $500k$ annotated images and covers 3 categories which consist of different subcategories: Vehicle (car, van, bus and truck), Cyclist (cyclist, motorcyclist and tricyclist ) and Pedestrian. We selected two scenes that were not included in the training set as validation data and used the mATE and mAOE of obstacles as the evaluation metrics.

\begin{table*}
\begin{center}
\tabcolsep=0.0075\linewidth
\begin{tabular}{l|c|ccccc|ccccc|c}
\hline
\multirow{2}*{Method} & \multirow{2}*{Reference} & \multicolumn{5}{c|}{Easy} & \multicolumn{5}{c|}{Hard} & Avg.  \\ \cline{3-12}
 & & NDS & mAP & mATE & mASE & mAOE & NDS & mAP & mATE & mASE & mAOE & NDS \\
\hline
\hline

SOLOFusion \cite{Park2022TimeWT} & ICLR 2023 & 0.308 & 0.129 & 0.878 & 0.144 & 0.517 & 0.202 & 0.066 & 0.844 & 0.144 & 1.000 & 0.255 \\
BEVDet4D \cite{huang2022bevdet4d} & arXiv & 0.428 & 0.200 & 0.896 & 0.094 & 0.041 & 0.393 & 0.139 & 0.922 & 0.099 & 0.038 & 0.411 \\
BEVDet \cite{huang2021bevdet} & arXiv & 0.506 & 0.299 & 0.742 & 0.079 & 0.042 & 0.445 & 0.184 & 0.754 & 0.087 & 0.043 & 0.476 \\
StreamPETR \cite{wang2023exploring} & ICCV 2023 & 0.619 & 0.513 & 0.690 & 0.102 & 0.032 & 0.496 & 0.284 & 0.739 & 0.107 & 0.031 & 0.558 \\
PETRv2 \cite{liu2023petrv2} & ICCV 2023 & 0.674 & 0.587 & 0.590 & 0.090 & 0.032 & 0.580 & 0.414 & 0.633 & 0.100 & 0.029 & 0.627 \\
BEVFormer \cite{li2022bevformer} & ECCV 2022 & 0.693 & 0.609 & 0.560 & 0.078 & 0.030 & 0.597 & 0.433 & 0.600 & 0.090 & 0.029 & 0.645 \\
DETR3D \cite{detr3d} & CoRL 2021 & 0.722 & 0.644 & 0.501 & 0.067 & 0.031 & 0.633 & 0.471 & 0.508 & 0.080 & 0.028 & 0.678 \\
RoBEV \cite{zhu2024rosceneslargescalemultiview3d}& ECCV 2024 & 0.753 & 0.684 & 0.442 & 0.058 & 0.031 & 0.672 & 0.524 & 0.438 & \textbf{0.077} & 0.027 & 0.713 \\
\textbf{RopeBEV (Ours)} & - & \textbf{0.786} & \textbf{0.721} & \textbf{0.435} & \textbf{0.056} & \textbf{0.030} & \textbf{0.685} & \textbf{0.545} & \textbf{0.416} & 0.078 & \textbf{0.027} & \textbf{0.736} \\
\hline
\end{tabular}
\end{center}
\caption{\textbf{Performance comparison of BEV methods on RoScenes val dataset.}}
\label{table_1}
\end{table*}

\subsection{Implementation Details}
For both the two datasets, the input image size is $960 \times 544$ and the backbone for image feature extraction is ResNet-50 \cite{he2016deep}. All models are trained for 12 epoches on a machine with 8 NVIDIA A100 GPUs, batch size = 2. We use AdamW \cite{loshchilov2017decoupled} as an optimzer with a cosine annealing learning rate schedule where the initial learning rate is set to $2e-4$. For RoScenes, the BEV grid size is set to $500 \times 500$, with a perception range defined as $X_{range} = [-160.0m, 160.0m]$ and $Y_{range} = [-20m, 800m]$. The origin of the BEV coordinate system is typically set to one of the cameras positioned at the edge of the perception area, oriented towards the area. In the case of the private dataset, the BEV grid is $300 \times 300$, with a perception range defined as $X_{range} = [-170.0m, 130.0m]$ and $Y_{range} = [-80.0m, 220.0m]$. Here, the origin of the BEV coordinate system is typically set to one of the cameras facing the intersection.

\subsection{Main Results}
Table \ref{table_1} illustrates the performance comparison on RoScenes. We compare our RopeBEV with the state-of-the-art multi-camera based methods, including BEVDet \cite{huang2021bevdet}, BEVDet4D \cite{huang2022bevdet4d}, SOLOFusion \cite{Park2022TimeWT}, BEVFormer \cite{li2022bevformer}, DETR3D \cite{detr3d}, PETRv2 \cite{liu2023petrv2}, StreamPETR \cite{wang2023exploring} and RoBEV \cite{zhu2024rosceneslargescalemultiview3d}. The results demonstrate that RopeBEV outperforms state-of-the-art methods by 0.023 in NDS. Compared to the baseline BEVFormer, RopeBEV achieved a significant improvement of 0.091 in NDS, demonstrating the effectiveness of our proposed enhancements.

\subsection{Ablation Study}

\begin{table}
\begin{center}
\small
\tabcolsep=0.17em
\begin{tabular}{l|cc|cc|cc|cc}
\hline
\multirow{2}*{Method} & \multicolumn{2}{c|}{Single-Single} & \multicolumn{2}{c|}{Single-All} & \multicolumn{2}{c|}{All-Single} & \multicolumn{2}{c}{All-All} \\
\cline{2-9}
 & Easy & Hard & Easy & Hard & Easy & Hard & Easy & Hard \\
\hline
\hline
DETR3D & 0.660 & 0.545 & 0.382 & 0.375 & 0.701 & 0.614 & 0.722 & 0.633 \\
PETRv2 & 0.649 & 0.512 & 0.376 & 0.359 & 0.636 & 0.563 & 0.674 & 0.580 \\
RoBEV & 0.683 & 0.571 & 0.396 & 0.387 & 0.720 & 0.631 & 0.753 & 0.672 \\
\textbf{RopeBEV} & \textbf{0.688} & \textbf{0.617} & \textbf{0.431} & \textbf{0.425} & \textbf{0.723} & \textbf{0.659} & \textbf{0.786} & \textbf{0.685} \\
\hline
\end{tabular}
\end{center}
\caption{\textbf{Transferability validation between a single scene \#001 and all scenes.} Single means \#001 and All means the whole scene. Single-All means training on the \#001 and validating on all scenes.}
\label{cross}
\end{table}

\textbf{Impact of cross-scene training.} 
For roadside perception tasks, generalization to new scenes is critical. In Table \ref{cross}, we report the cross-scene validation results based on RoScenes. When training and validating only on Scene \#001, all methods show strong performance. This is because the camera layout in Scene \#001 is fixed. However, when using Scene \#001 for training but evaluating across all scenes, the performance of the compared methods drops significantly, while RopeBEV maintains high accuracy. This demonstrates the effectiveness of BEV data augmentation in enhancing feature extractor capabilities. In the All-Single experiment group, it's evident that training with full data further improves performance on the single-scene evaluation set. This is logical, as the increase in dataset size and the greater diversity in camera layout effectively enhance the model's generalization capabilities.

\begin{table}
\begin{center}
\small
\tabcolsep=0.2em
\begin{tabular}{l|ccc|ccc}
\hline
\multirow{2}*{Method} & \multicolumn{3}{c|}{mATE} & \multicolumn{3}{c}{mAOE} \\
\cline{2-7}
 & Vehicle & Cyclist & Pedestrian & Vehicle & Cyclist & Pedestrian \\
\hline
\hline
w/o  & 0.639 & 0.500 & 0.523 & 0.030 & 0.149 & 0.752 \\
w/ & 0.633 & 0.507 & 0.526 & 0.031 & 0.121 & 0.629 \\
\hline
\end{tabular}
\end{center}
\caption{\textbf{Ablation Study of camera rotation embedding on private dataset.} w/o means without camera rotation embedding, while w/ means using camera rotation embedding.}
\label{orientation}
\end{table}

\textbf{Impact of camera rotation embedding.} 
The RoScenes dataset includes only motor vehicle categories, while the issue of orientation ambiguity is more pronounced in smaller objects such as pedestrian and cyclist. Therefore, we conducted a validation of the impact of camera rotation embedding using our private dataset. As shown in Table \ref{orientation}, after incorporating the camera rotation embedding, the orientation angle of pedestrian and cyclist was significantly improved, with the mAOE reduced by 0.123 and 0.028 for each category. Additionally, the camera rotation embedding had no impact on the mATE of the obstacles.

\subsection{Qualitative Results}
\begin{figure*}[ht]
\centering
\includegraphics[width=170mm]{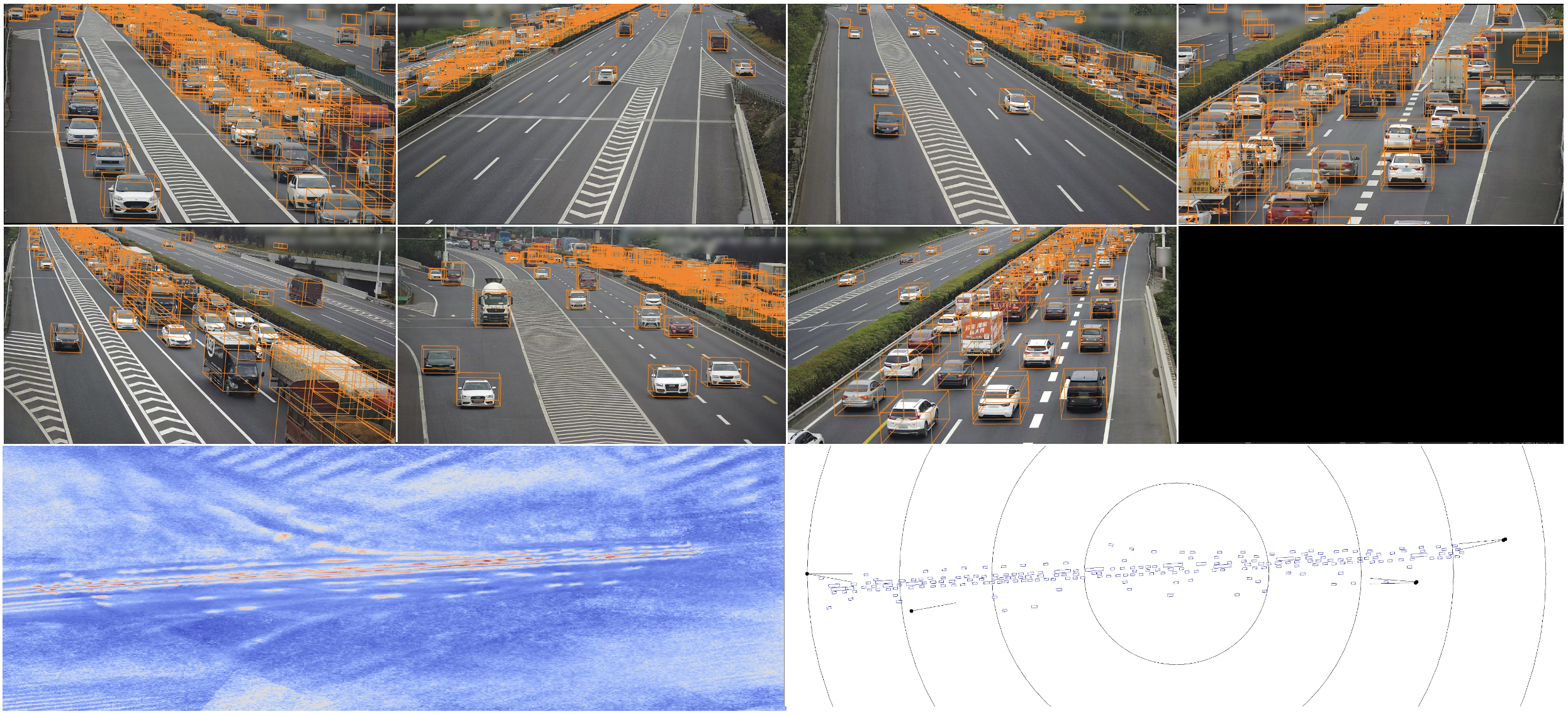}
\caption{\textbf{Visualization results of our proposed RopeBEV on Roscenes.} The first two rows are images captured by pinhole cameras. The last row presents the visualization of the BEV Feature on the left and the detection results from the BEV perspective on the right. The radial spacing of the equidistant circles is 100 meters.}
\label{vis}
\end{figure*}

\begin{figure}[ht]
\begin{center}
\includegraphics[width=1.0\linewidth]{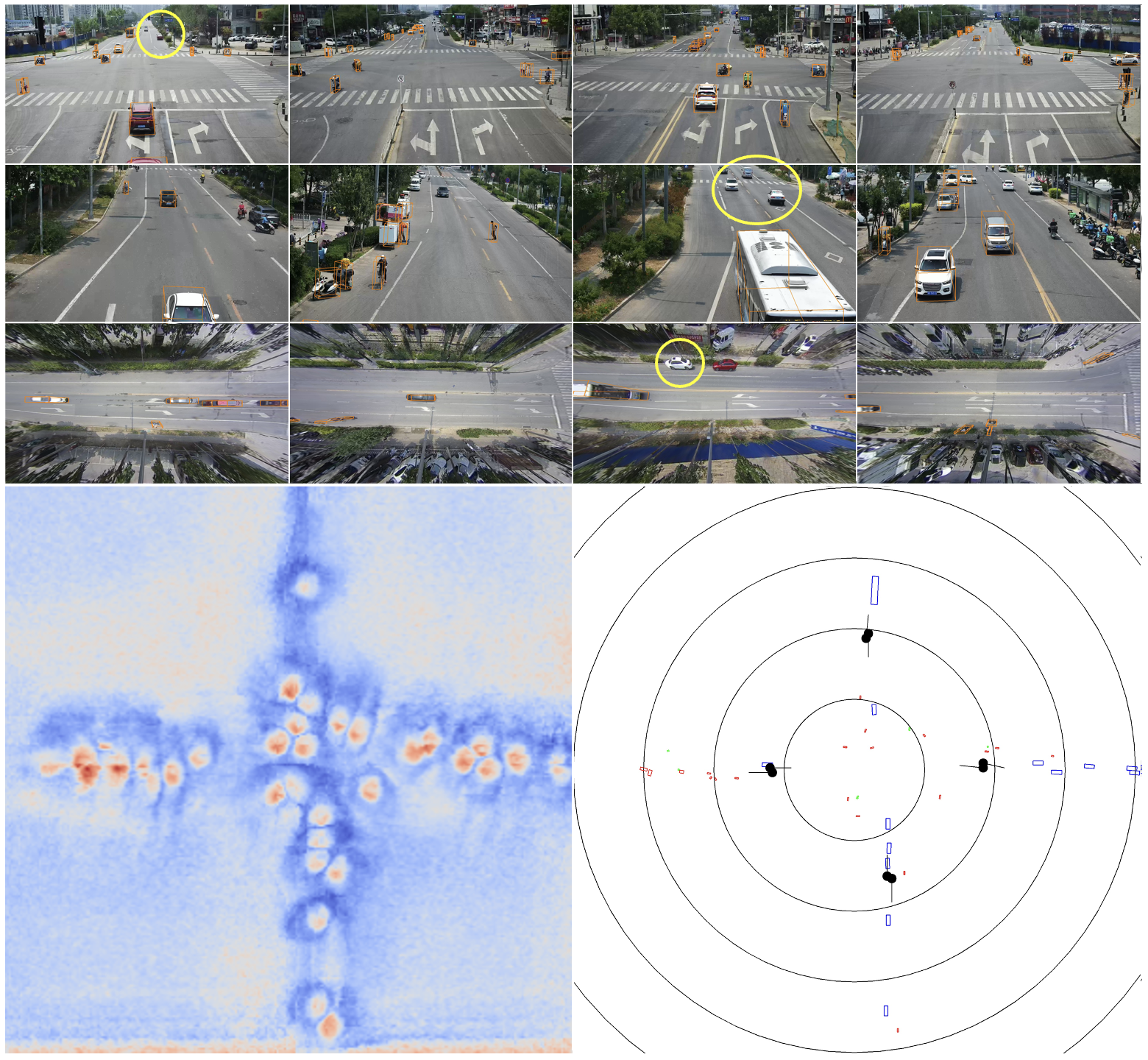}
\end{center}
\caption{\textbf{Visualization results of our proposed RopeBEV on Private Dataset.} The first two rows are images captured by pinhole cameras, the third row shows images captured by fisheye cameras, and the fourth row presents the visualization of the BEV Feature on the left and the detection results from the BEV perspective on the right. The yellow circles indicate objects in the exit directions filtered by ROIMask. The radial spacing of the equidistant circles is 30 meters.}
\label{vis_private}
\end{figure}

In Figures \ref{vis} and \ref{vis_private}, we provide visualizations of RopeBEV's results on the Roscenes and private datasets, respectively. During inference on the Roscenes dataset, CamMask was applied, as shown by the black region in the bottom right corner of Figure \ref{vis}, where one camera is missing compared to the maximum supported number. For the private dataset, we utilized ROIMask during inference (with ROIMask drawn based on the rules from the Rope3D \cite{ye2022rope3d} dataset), where the exit directions of the intersections were excluded from BEV space modeling and thus had no perception results (as indicated by the yellow circle in Figure \ref{vis_private}).

\section{Conclusion}
This paper systematically analyzes the differences between roadside and vehicle-side multi-camera perception scenarios, including the diversity in camera poses, the uncertainty in camera numbers, the sparsity in perception regions, and the ambiguity in orientation angles. To address these challenges, we propose RopeBEV, the first dense BEV method for roadside multi-camera perception. RopeBEV introduces BEV augmentation to resolve the imbalance in training learnable queries caused by the diverse poses of roadside cameras, employs CamMask and ROIMask to support varying numbers of cameras and customizable perception regions, and addresses the orientation angle ambiguity through camera rotation embedding. Experiments on the RoScenes and private datasets demonstrate the superior roadside perception performance of RopeBEV.

\section{Limitations and Future Work}
Although RopeBEV is designed to support lane segmen-
tation and scene mapping tasks, we are currently unable to evaluate its performance due to the lack of roadside multi-camera dataset for these tasks. Additionally, due to physical constraints, the time stamp differences between roadside multi-camera systems are generally larger than those in vehicle-side systems, which can result in non-aligned information within the input framse, thus affecting perception outcomes. Explicitly incorporating temporal embedding or feature alignment across time dimensions is one of the future directions we intend to pursue. Furthermore, we aim to expand RopeBEV to support tracking and prediction tasks, and explore the integration of additional data modalities, such as lidar or radar, to achieve a multimodal roadside BEV approach.

{\small
\bibliographystyle{ieee_fullname}
\bibliography{main}

\begin{thebibliography}{10}\itemsep=-1pt

\bibitem{Borse_2023_WACV}
Shubhankar Borse, Marvin Klingner, Varun~Ravi Kumar, Hong Cai, Abdulaziz Almuzairee, Senthil Yogamani, and Fatih Porikli.
\newblock X-align: Cross-modal cross-view alignment for bird's-eye-view segmentation.
\newblock In {\em Proceedings of the IEEE/CVF Winter Conference on Applications of Computer Vision (WACV)}, pages 3287--3297, January 2023.

\bibitem{Chen_2024_WACV}
Qiuxiao Chen and Xiaojun Qi.
\newblock Residual graph convolutional network for bird's-eye-view semantic segmentation.
\newblock In {\em Proceedings of the IEEE/CVF Winter Conference on Applications of Computer Vision (WACV)}, pages 3324--3331, January 2024.

\bibitem{Chen_2023_ICCV}
Ziming Chen, Yifeng Shi, and Jinrang Jia.
\newblock Transiff: An instance-level feature fusion framework for vehicle-infrastructure cooperative 3d detection with transformers.
\newblock In {\em Proceedings of the IEEE/CVF International Conference on Computer Vision (ICCV)}, pages 18205--18214, October 2023.

\bibitem{Fadadu_2022_WACV}
Sudeep Fadadu, Shreyash Pandey, Darshan Hegde, Yi Shi, Fang-Chieh Chou, Nemanja Djuric, and Carlos Vallespi-Gonzalez.
\newblock Multi-view fusion of sensor data for improved perception and prediction in autonomous driving.
\newblock In {\em Proceedings of the IEEE/CVF Winter Conference on Applications of Computer Vision (WACV)}, pages 2349--2357, January 2022.

\bibitem{fan2021embracingsinglestride3d}
Lue Fan, Ziqi Pang, Tianyuan Zhang, Yu-Xiong Wang, Hang Zhao, Feng Wang, Naiyan Wang, and Zhaoxiang Zhang.
\newblock Embracing single stride 3d object detector with sparse transformer, 2021.

\bibitem{fan2023calibrationfreebevrepresentationinfrastructure}
Siqi Fan, Zhe Wang, Xiaoliang Huo, Yan Wang, and Jingjing Liu.
\newblock Calibration-free bev representation for infrastructure perception, 2023.

\bibitem{hao2024rcooper}
Ruiyang Hao, Siqi Fan, Yingru Dai, Zhenlin Zhang, Chenxi Li, Yuntian Wang, Haibao Yu, Wenxian Yang, Yuan Jirui, and Zaiqing Nie.
\newblock Rcooper: A real-world large-scale dataset for roadside cooperative perception.
\newblock In {\em Proceedings of the IEEE/CVF Conference on Computer Vision and Pattern Recognition (CVPR)}, pages 22347--22357, 2024.

\bibitem{he2016deep}
Kaiming He, Xiangyu Zhang, Shaoqing Ren, and Jian Sun.
\newblock Deep residual learning for image recognition.
\newblock In {\em Proceedings of the IEEE conference on computer vision and pattern recognition}, pages 770--778, 2016.

\bibitem{Where2comm}
Yue Hu, Shaoheng Fang, Zixing Lei, Yiqi Zhong, and Siheng Chen.
\newblock Where2comm: Communication-efficient collaborative perception via spatial confidence maps.
\newblock In {\em Thirty-sixth Conference on Neural Information Processing Systems (Neurips)}, November 2022.

\bibitem{huang2022bevdet4d}
Junjie Huang and Guan Huang.
\newblock Bevdet4d: Exploit temporal cues in multi-camera 3d object detection.
\newblock {\em arXiv preprint arXiv:2203.17054}, 2022.

\bibitem{huang2021bevdet}
Junjie Huang, Guan Huang, Zheng Zhu, Ye Yun, and Dalong Du.
\newblock Bevdet: High-performance multi-camera 3d object detection in bird-eye-view.
\newblock {\em arXiv preprint arXiv:2112.11790}, 2021.

\bibitem{jia2023monouni}
Jinrang Jia, Zhenjia Li, and Yifeng Shi.
\newblock Monouni: A unified vehicle and infrastructure-side monocular 3d object detection network with sufficient depth clues.
\newblock In {\em Thirty-seventh Conference on Neural Information Processing Systems}, 2023.

\bibitem{nwad121}
Jinrang Jia, Yifeng Shi, Yuli Qu, Rui Wang, Xing Xu, and Hai Zhang.
\newblock {Competition for roadside camera monocular 3D object detection}.
\newblock {\em National Science Review}, 10(6):nwad121, 05 2023.

\bibitem{danet}
Bo Ju, Wei Yang, Jinrang Jia, Xiaoqing Ye, Qu Chen, Xiao Tan, Hao Sun, Yifeng Shi, and Errui Ding.
\newblock Danet: Dimension apart network for radar object detection.
\newblock In {\em Proceedings of the 2021 International Conference on Multimedia Retrieval}, ICMR '21, page 533–539, New York, NY, USA, 2021. Association for Computing Machinery.

\bibitem{DUSA_2023_ACMMM}
Xianghao Kong, Wentao Jiang, Jinrang Jia, Yifeng Shi, Runsheng Xu, and Si Liu.
\newblock Dusa: Decoupled unsupervised sim2real adaptation for vehicle-to-everything collaborative perception.
\newblock In {\em Proceedings of the 31st ACM International Conference on Multimedia}, October 2023.

\bibitem{li2022bevdepth}
Yinhao Li, Zheng Ge, Guanyi Yu, Jinrong Yang, Zengran Wang, Yukang Shi, Jianjian Sun, and Zeming Li.
\newblock Bevdepth: Acquisition of reliable depth for multi-view 3d object detection.
\newblock {\em arXiv preprint arXiv:2206.10092}, 2022.

\bibitem{li2023fast}
Yangguang Li, Bin Huang, Zeren Chen, Yufeng Cui, Feng Liang, Mingzhu Shen, Fenggang Liu, Enze Xie, Lu Sheng, Wanli Ouyang, et~al.
\newblock Fast-bev: A fast and strong bird's-eye view perception baseline.
\newblock {\em arXiv preprint arXiv:2301.12511}, 2023.

\bibitem{10550788}
Zhenjia Li, Jinrang Jia, and Yifeng Shi.
\newblock Monolss: Learnable sample selection for monocular 3d detection.
\newblock In {\em 2024 International Conference on 3D Vision (3DV)}, pages 1125--1135, 2024.

\bibitem{li2022bevformer}
Zhiqi Li, Wenhai Wang, Hongyang Li, Enze Xie, Chonghao Sima, Tong Lu, Yu Qiao, and Jifeng Dai.
\newblock Bevformer: Learning bird’s-eye-view representation from multi-camera images via spatiotemporal transformers.
\newblock {\em arXiv preprint arXiv:2203.17270}, 2022.

\bibitem{Sparse4D}
Xuewu Lin, Tianwei Lin, Zixiang Pei, Lichao Huang, and Zhizhong Su.
\newblock Sparse4d: Multi-view 3d object detection with sparse spatial-temporal fusion, 2022.

\bibitem{lin2023sparse4d}
Xuewu Lin, Tianwei Lin, Zixiang Pei, Lichao Huang, and Zhizhong Su.
\newblock Sparse4d v2: Recurrent temporal fusion with sparse model.
\newblock {\em arXiv preprint arXiv:2305.14018}, 2023.

\bibitem{liu2023sparsebevhighperformancesparse3d}
Haisong Liu, Yao Teng, Tao Lu, Haiguang Wang, and Limin Wang.
\newblock Sparsebev: High-performance sparse 3d object detection from multi-camera videos, 2023.

\bibitem{liu2022petr}
Yingfei Liu, Tiancai Wang, Xiangyu Zhang, and Jian Sun.
\newblock Petr: Position embedding transformation for multi-view 3d object detection.
\newblock {\em arXiv preprint arXiv:2203.05625}, 2022.

\bibitem{liu2023petrv2}
Yingfei Liu, Junjie Yan, Fan Jia, Shuailin Li, Aqi Gao, Tiancai Wang, and Xiangyu Zhang.
\newblock Petrv2: A unified framework for 3d perception from multi-camera images.
\newblock In {\em Proceedings of the IEEE/CVF International Conference on Computer Vision}, pages 3262--3272, 2023.

\bibitem{loshchilov2017decoupled}
I Loshchilov.
\newblock Decoupled weight decay regularization.
\newblock {\em arXiv preprint arXiv:1711.05101}, 2017.

\bibitem{lu2023robust}
Yifan Lu, Quanhao Li, Baoan Liu, Mehrdad Dianati, Chen Feng, Siheng Chen, and Yanfeng Wang.
\newblock Robust collaborative 3d object detection in presence of pose errors.
\newblock In {\em 2023 IEEE International Conference on Robotics and Automation (ICRA)}, pages 4812--4818. IEEE, 2023.

\bibitem{Park2022TimeWT}
Jinhyung Park, Chenfeng Xu, Shijia Yang, Kurt Keutzer, Kris Kitani, Masayoshi Tomizuka, and Wei Zhan.
\newblock Time will tell: New outlooks and a baseline for temporal multi-view 3d object detection.
\newblock 2023.

\bibitem{Peng_2023_WACV}
Lang Peng, Zhirong Chen, Zhangjie Fu, Pengpeng Liang, and Erkang Cheng.
\newblock Bevsegformer: Bird's eye view semantic segmentation from arbitrary camera rigs.
\newblock In {\em Proceedings of the IEEE/CVF Winter Conference on Applications of Computer Vision (WACV)}, pages 5935--5943, January 2023.

\bibitem{peng2022did}
Liang Peng, Xiaopei Wu, Zheng Yang, Haifeng Liu, and Deng Cai.
\newblock Did-m3d: Decoupling instance depth for monocular 3d object detection.
\newblock In {\em European Conference on Computer Vision}, 2022.

\bibitem{philion2020lift}
Jonah Philion and Sanja Fidler.
\newblock Lift, splat, shoot: Encoding images from arbitrary camera rigs by implicitly unprojecting to 3d.
\newblock In {\em Proceedings of the European Conference on Computer Vision}, 2020.

\bibitem{Qiao_2023_WACV}
Donghao Qiao and Farhana Zulkernine.
\newblock Adaptive feature fusion for cooperative perception using lidar point clouds.
\newblock In {\em Proceedings of the IEEE/CVF Winter Conference on Applications of Computer Vision (WACV)}, pages 1186--1195, January 2023.

\bibitem{Rukhovich_2022_WACV}
Danila Rukhovich, Anna Vorontsova, and Anton Konushin.
\newblock Imvoxelnet: Image to voxels projection for monocular and multi-view general-purpose 3d object detection.
\newblock In {\em Proceedings of the IEEE/CVF Winter Conference on Applications of Computer Vision (WACV)}, pages 2397--2406, January 2022.

\bibitem{vaswani2023attentionneed}
Ashish Vaswani, Noam Shazeer, Niki Parmar, Jakob Uszkoreit, Llion Jones, Aidan~N. Gomez, Lukasz Kaiser, and Illia Polosukhin.
\newblock Attention is all you need, 2023.

\bibitem{wang2023exploring}
Shihao Wang, Yingfei Liu, Tiancai Wang, Ying Li, and Xiangyu Zhang.
\newblock Exploring object-centric temporal modeling for efficient multi-view 3d object detection.
\newblock {\em arXiv preprint arXiv:2303.11926}, 2023.

\bibitem{wang2024bevspreadspreadvoxelpooling}
Wenjie Wang, Yehao Lu, Guangcong Zheng, Shuigen Zhan, Xiaoqing Ye, Zichang Tan, Jingdong Wang, Gaoang Wang, and Xi Li.
\newblock Bevspread: Spread voxel pooling for bird's-eye-view representation in vision-based roadside 3d object detection, 2024.

\bibitem{detr3d}
Yue Wang, Vitor Guizilini, Tianyuan Zhang, Yilun Wang, Hang Zhao, , and Justin~M. Solomon.
\newblock Detr3d: 3d object detection from multi-view images via 3d-to-2d queries.
\newblock In {\em The Conference on Robot Learning ({CoRL})}, 2021.

\bibitem{xu2022v2xvit}
Runsheng Xu, Hao Xiang, Zhengzhong Tu, Xin Xia, Ming-Hsuan Yang, and Jiaqi Ma.
\newblock V2x-vit: Vehicle-to-everything cooperative perception with vision transformer.
\newblock In {\em Proceedings of the European Conference on Computer Vision (ECCV)}, 2022.

\bibitem{yang2023bevformer}
Chenyu Yang, Yuntao Chen, Hao Tian, Chenxin Tao, Xizhou Zhu, Zhaoxiang Zhang, Gao Huang, Hongyang Li, Yu Qiao, Lewei Lu, et~al.
\newblock Bevformer v2: Adapting modern image backbones to bird's-eye-view recognition via perspective supervision.
\newblock In {\em Proceedings of the IEEE/CVF Conference on Computer Vision and Pattern Recognition}, pages 17830--17839, 2023.

\bibitem{Yang2022BEVFormerVA}
Chenyu Yang, Yuntao Chen, Haofei Tian, Chenxin Tao, Xizhou Zhu, Zhaoxiang Zhang, Gao Huang, Hongyang Li, Y. Qiao, Lewei Lu, Jie Zhou, and Jifeng Dai.
\newblock Bevformer v2: Adapting modern image backbones to bird's-eye-view recognition via perspective supervision.
\newblock {\em ArXiv}, 2022.

\bibitem{yang2023bevheightrobustvisualcentric}
Lei Yang, Tao Tang, Jun Li, Peng Chen, Kun Yuan, Li Wang, Yi Huang, Xinyu Zhang, and Kaicheng Yu.
\newblock Bevheight++: Toward robust visual centric 3d object detection, 2023.

\bibitem{yang2023monogaeroadsidemonocular3d}
Lei Yang, Jiaxin Yu, Xinyu Zhang, Jun Li, Li Wang, Yi Huang, Chuang Zhang, Hong Wang, and Yiming Li.
\newblock Monogae: Roadside monocular 3d object detection with ground-aware embeddings, 2023.

\bibitem{yang2023bevheight}
Lei Yang, Kaicheng Yu, Tao Tang, Jun Li, Kun Yuan, Li Wang, Xinyu Zhang, and Peng Chen.
\newblock Bevheight: A robust framework for vision-based roadside 3d object detection.
\newblock In {\em IEEE/CVF Conf.~on Computer Vision and Pattern Recognition (CVPR)}, Mar. 2023.

\bibitem{ye2022rope3d}
Xiaoqing Ye, Mao Shu, Hanyu Li, Yifeng Shi, Yingying Li, Guangjie Wang, Xiao Tan, and Errui Ding.
\newblock Rope3d: The roadside perception dataset for autonomous driving and monocular 3d object detection task.
\newblock In {\em Proceedings of the IEEE/CVF Conference on Computer Vision and Pattern Recognition}, pages 21341--21350, 2022.

\bibitem{Yu_2022_CVPR}
Haibao Yu, Yizhen Luo, Mao Shu, Yiyi Huo, Zebang Yang, Yifeng Shi, Zhenglong Guo, Hanyu Li, Xing Hu, Jirui Yuan, and Zaiqing Nie.
\newblock Dair-v2x: A large-scale dataset for vehicle-infrastructure cooperative 3d object detection.
\newblock In {\em Proceedings of the IEEE/CVF Conference on Computer Vision and Pattern Recognition (CVPR)}, pages 21361--21370, June 2022.

\bibitem{v2x-seq}
Haibao Yu, Wenxian Yang, Hongzhi Ruan, Zhenwei Yang, Yingjuan Tang, Xu Gao, Xin Hao, Yifeng Shi, Yifeng Pan, Ning Sun, Juan Song, Jirui Yuan, Ping Luo, and Zaiqing Nie.
\newblock V2x-seq: A large-scale sequential dataset for vehicle-infrastructure cooperative perception and forecasting.
\newblock In {\em Proceedings of the IEEE/CVF Conference on Computer Vision and Pattern Recognition}, 2023.

\bibitem{Zhang_2023_ICCV}
Yunpeng Zhang, Zheng Zhu, and Dalong Du.
\newblock Occformer: Dual-path transformer for vision-based 3d semantic occupancy prediction.
\newblock In {\em Proceedings of the IEEE/CVF International Conference on Computer Vision (ICCV)}, pages 9433--9443, October 2023.

\bibitem{zhu2024rosceneslargescalemultiview3d}
Xiaosu Zhu, Hualian Sheng, Sijia Cai, Bing Deng, Shaopeng Yang, Qiao Liang, Ken Chen, Lianli Gao, Jingkuan Song, and Jieping Ye.
\newblock Roscenes: A large-scale multi-view 3d dataset for roadside perception, 2024.

\bibitem{10186723}
Walter Zimmer, Joseph Birkner, Marcel Brucker, Huu Tung~Nguyen, Stefan Petrovski, Bohan Wang, and Alois~C. Knoll.
\newblock Infradet3d: Multi-modal 3d object detection based on roadside infrastructure camera and lidar sensors.
\newblock In {\em 2023 IEEE Intelligent Vehicles Symposium (IV)}, pages 1--8, 2023.

\end{thebibliography}
}

\end{document}